\pdfoutput=1

\documentclass[11pt]{article}

\usepackage[]{ACL2023}

\usepackage{times}
\usepackage{latexsym}
\usepackage{booktabs}
\usepackage{graphicx}
\usepackage{rotating}

\usepackage[T1]{fontenc}

\usepackage[utf8]{inputenc}

\usepackage{microtype}

\usepackage{inconsolata}

\usepackage{amsmath}

\newcommand*\samethanks[1][\value{footnote}]{\footnotemark[#1]}

\title{Large Language Models as Sous Chefs: Revising Recipes with GPT-3}

\author{Alyssa Hwang\thanks{\hspace{4pt} Equal contribution} \and Bryan Li\samethanks \and Zhaoyi Hou\samethanks \and Dan Roth \\
        Department of Computer and Information Science \\
        University of Pennsylvania \\
        \{\texttt{ahwang16}, \texttt{bryanli}, \texttt{joeyhou}, \texttt{danroth}\}\texttt{@seas.upenn.edu}}

\begin{document}
\maketitle
\begin{abstract}
With their remarkably improved text generation and prompting capabilities, large language models can adapt existing written information into forms that are easier to use and understand. In our work, we focus on recipes as an example of complex, diverse, and widely used instructions. 
We develop a prompt grounded in the original recipe and ingredients list that breaks recipes down into simpler steps. We apply this prompt to recipes from various world cuisines, and experiment with several large language models (LLMs), finding best results with GPT-3.5. We also contribute an Amazon Mechanical Turk task that is carefully designed to reduce fatigue while collecting human judgment of the quality of recipe revisions. We find that annotators usually prefer the revision over the original, demonstrating a promising application of LLMs in serving as digital \textit{sous chefs} for recipes and beyond. We release our prompt, code\footnote{\url{https://github.com/manestay/project2-complex-tasks}}, and MTurk template\footnote{\url{https://github.com/ahwang16/mturk-templates/tree/master/compare_recipes}} for public use.
\end{abstract}

\section{Introduction}
We have access to vast amounts of written information, but many of these resources are written in different styles, levels of complexity, and domains. Previous work in adapting information to be easier to use and understand have relied on advances in general knowledge, event reasoning, text simplification, and other subtasks in Natural Language Processing. Today, the tremendous improvement in large language models, especially their zero-shot prompting and in-context learning capabilities, has made it easier than ever to leverage NLP research for user-centered goals.

Our work focuses on revising food recipes as a key use case for adapting written information. Recipes are abundant, diverse, and widely used, making them a prime candidate to explore how well GPT-3.5~\cite{brown2020language}---one of the currently most powerful LLMs---can handle them. For each recipe sampled across various world cuisines, we ask GPT-3.5 to use the original recipe and ingredients list to produce an adaptation that is easier to follow.

We rely on human judgment collected through a carefully designed task on Amazon Mechanical Turk to evaluate revised recipes. Instead of displaying original and revised recipes side by side and asking annotators to evaluate many parts of both at once, we simplify the process by displaying one step at a time. Beyond general usability, we ask annotators to identify new, missing, or incorrect information in a revised recipe. GPT-3.5 often introduces new information to clarify implicit information. This new information can be classified as either a `hallucination', which is detrimental to generation quality, or an `elaboration', which is helpful for end-users.

Our work makes these three key contributions:
\begin{enumerate}
    \item We propose the task of using LLMs to improve the presentation of complex, written information. Specifically, we prompt an LLM to revise a given recipe.
    \item We describe our experimentation to design a successful prompt. We find that by including grounding information (here the original recipe and list of ingredients) in the prompt, LLM generations will emphasize helpful elaborations over detrimental hallucinations.
    \item We design a human evaluation task that enables annotators to make granular, step-by-step comparisons between the original and revised recipes. This granularity both reduces the mental strain of annotation and allows for step-level analysis of the model's revisions. We find that annotators prefer revised recipes.
\end{enumerate}

\section{Methods}

\subsection{Data}

We consider the  Recipes1M+ dataset \citep{marin2019learning, salvador2017learning}, which consists of over a million multimodal recipes collected from across the internet. For the purposes of this work, we use stratified random sampling to select a 100-recipe subset. First, we manually select 10 Recipes1M+ \texttt{class} attributes (which are frequent ngrams occurring in recipe titles): \texttt{[couscous salad, apple pie, clam chowder, lentil soup, chicken enchiladas, beef stroganoff, coconut macaroons, chow mein, pad thai, paella]}. These attributes are diverse with respect to world cuisine, course, and preparation methods. For each attribute, we then sample 5 long recipes (11-16 steps) and 5 short recipes (5-10 steps). This follows from the length distribution of the full Recipes1M+ dataset, which peaks at 10 steps. The resultant subset consists of 100 recipes.

\subsection{Large Language Models Used}
We start with two models for recipe revision: \texttt{text-davinci-002} and \texttt{text-davinci-003}.\footnote{\url{https://platform.openai.com/playground}} After a manual evaluation of several pilot generations, we move forward with \texttt{text-davinci-003}. Anecdotally, we observe that \texttt{text-davinci-003} writes slightly more informative steps and follows prompt instructions better.

\subsection{Prompting Approach}
We now present our approach to prompting an LLM to generate clear, executable recipes. We first cover our final, most successful prompt. We then discuss insights from prior experimentation that led to it.

\begin{figure}[ht]
    \centering
    \small
    \noindent\fbox{%
    \parbox{0.95\linewidth}{%
        You are tasked with revising a food recipe, given a set of ingredients and the original steps. Make sure each generated step is executable (example: do not say "Enjoy!") and not too complex. Ensure that your steps still make the same dish as the original steps (example: for a salad dish, no need to bake it). \\
        Valid operations:\\
        A) Keep a step the same\\
        B) Delete a unneeded step\\
        C) Revise a unclear step\\
        D) Split a complex step into multiple steps\\
        E) Add a step to improve clarity\\
        F) Combine two steps which go together\\ \\        
        Here are demonstrations, formatted as "[input]" -> "[output]":\\
        A) "Simmer for 30 minutes" -> "Simmer for 30 minutes"\\
        B) "Enjoy!" -> [null (always)]\\
        C) "Add the veggies" -> "Add in the peas and mushrooms"\\
        D) "Knead the dough, and let rise overnight" -> "Knead the dough" + "Let it rise overnight"\\
        E) [null (when implied)] -> "Preheat the oven to 350 F"\\
        F) "Stir in potatoes" + "Stir in carrots" -> "Stir in potatoes and carrots"\\
        -{}-{}-{} \\
        \textsf{[title]}
        
        -{}-{}-{} \newline
         \textcolor{BrickRed}{\textsf{\noindent 
         * Water \\ ...}}\\
        -{}-{}-{}\\
        Original Recipe\\
         \textcolor{DarkOrchid}{\textsf{\noindent 
         1. Boil the water. \\ ...}}\\
        -{}-{}-{}\\
        Revised Recipe\\
        1. 
            }%
        }
    \caption{The prompt used to elicit an LLM to revise a recipe. \textsf{san serif} texts are filled in per recipe, while serif text is consistent across recipes. The text is color coded as: Instructions (black), \textcolor{BrickRed}{Ingredients}, \textcolor{DarkOrchid}{Original Recipe}. }
    \label{fig:revise_prompt}
\end{figure}
Our most successful prompt, depicted in Figure~\ref{fig:revise_prompt}, consists of three sections: instructions, a set of ingredients, and the original recipe.

\paragraph{Instructions} The instructions are verbose (163 words) to ensure the task is as well defined as possible. The list of ``Valid operations'' allows the model to make a diverse set of revisions, while the ``demonstrations'' clarify how texts are transformed. We find that separating the operations and demonstrations produced the best results.

We also add clarifications in parentheses to catch some problem cases, e.g. ``example: for a salad dish, no need to bake it'' (though there is still no guarantee against hallucinations).

\paragraph{Ingredients} Including the ingredients gives the model more information to base its revision decisions on. Ingredients are given in a bulleted list, which is generally between 50 and 100 words.

\paragraph{Original Recipe} 
The original steps are given in a numbered list and vary widely in terms of style, length, and formatting.

\subsection{Preliminary Prompting Experiments}
To arrive at this formulation, we experimented with several versions of the prompt and task.

\paragraph{Direct Generation}
Our initial task was to ask the LLM to \textbf{generate} a recipe with just the name of the dish. We thought this might work reasonably well, given that models such as GPT-3 have very likely seen many instances of food recipes during training. We provided a one-sentence instruction and the recipe title. We then experimented by eliciting completions for prompt configurations: comparing zero-shot and few-shot prompts and varying the presence of the ingredients list, for a total of four experimental settings. For the few-shot setting, each prompt uses two randomly sampled examples from the larger set of recipes for the dishes of interest. As found by \citet{patel2023bidirectional}, prompts with dynamically sampled examples allow for more diverse generations as opposed to static examples. 

We found that including ingredients greatly improved the faithfulness of the generated recipe to the dish. For example, when generating a recipe for ``low-fat and dairy-free clam chowder,'' the prompts without ingredients tended to use coconut milk (which is assuredly not low-fat). When the original ingredients were included, the generations would correctly use soy or rice milk. When including few-shot examples, we found that the examples mostly affected the style of the output text while the content seemed to be roughly the same.

We noticed that in all four configurations, the model struggled to account for modifications or qualifications and tended towards a more generic dish. For example, the generation for ``mini apple pies'' says to only make 1 large pie. Furthermore, many recipes may sound reasonable but actually make a different dish. The generation for ``paleo paella,'' for instance, says to use cauliflower \textit{florets} as a base, but paella is a rice dish, so the paleo variant should use \textit{riced} cauliflower. Such issues are especially problematic as they require domain knowledge which home cooks may lack, which would result them making dishes different from their initial intent.

A further complication when designing a human evaluation task for a given recipe (either human-written or machine-generated). We would like a task that allows the annotator to examine one step at a time. However, it is tough to say whether a step of a recipe is good or not given the open-ended, subjective nature of cooking.\footnote{This feedback was provided to us by researchers of our lab group.} Therefore, from the direct generation experiments, we learn that providing more information to the model allows for better quality generations. This motivates us to instead include the original recipe in the prompt and have a model revise it.

\paragraph{Generation + Identifying Operations} We tried to ask the model to additionally output the letter of the step used (A-F); however, this was  nearly always A and thus not faithful. We leave this issue to future work, noting that the letter could provide a trace to the model's reasoning and thus better end-user interpretability.

\paragraph{Why Not Fine-Tune?} While we considered fine-tuning, two reasons led to our decision to use prompting instead. First, there is no gold training data for revising recipes (and collecting it was prohibitively expensive and subjective). Second, interfaces to LLMs such as ChatGPT and Bard have been widely adopted by the public already. Therefore, we also use off-the-shelf LLMs so our findings can better align with how people actually use them.

\section{Human Evaluation}
\begin{sidewaysfigure*}
    \centering
    \includegraphics[angle=270, width=0.8\textwidth]{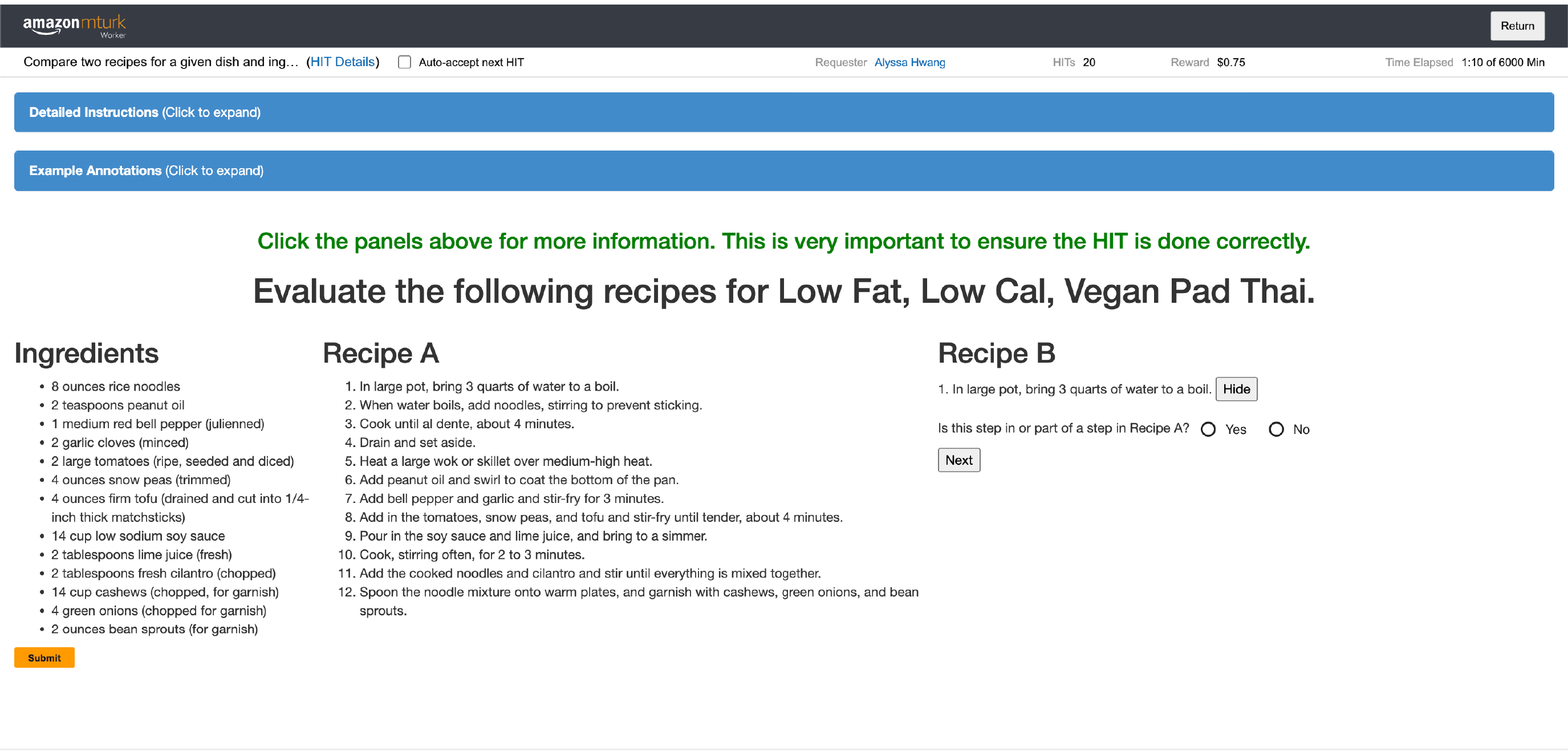}
    \caption{The Opening Page for Our MTurk Task. When a annotator first accepts the HIT, they will see the dish name, ingredients list, and one full recipe (Recipe A). They will answer questions about the second recipe (Recipe B) one step at a time. Detailed instructions and examples are available in prominently displayed collapsible banners above.}
    \label{fig:mturk_open}
\end{sidewaysfigure*}

\begin{figure}[ht]
    \centering
    \includegraphics[width=\columnwidth]{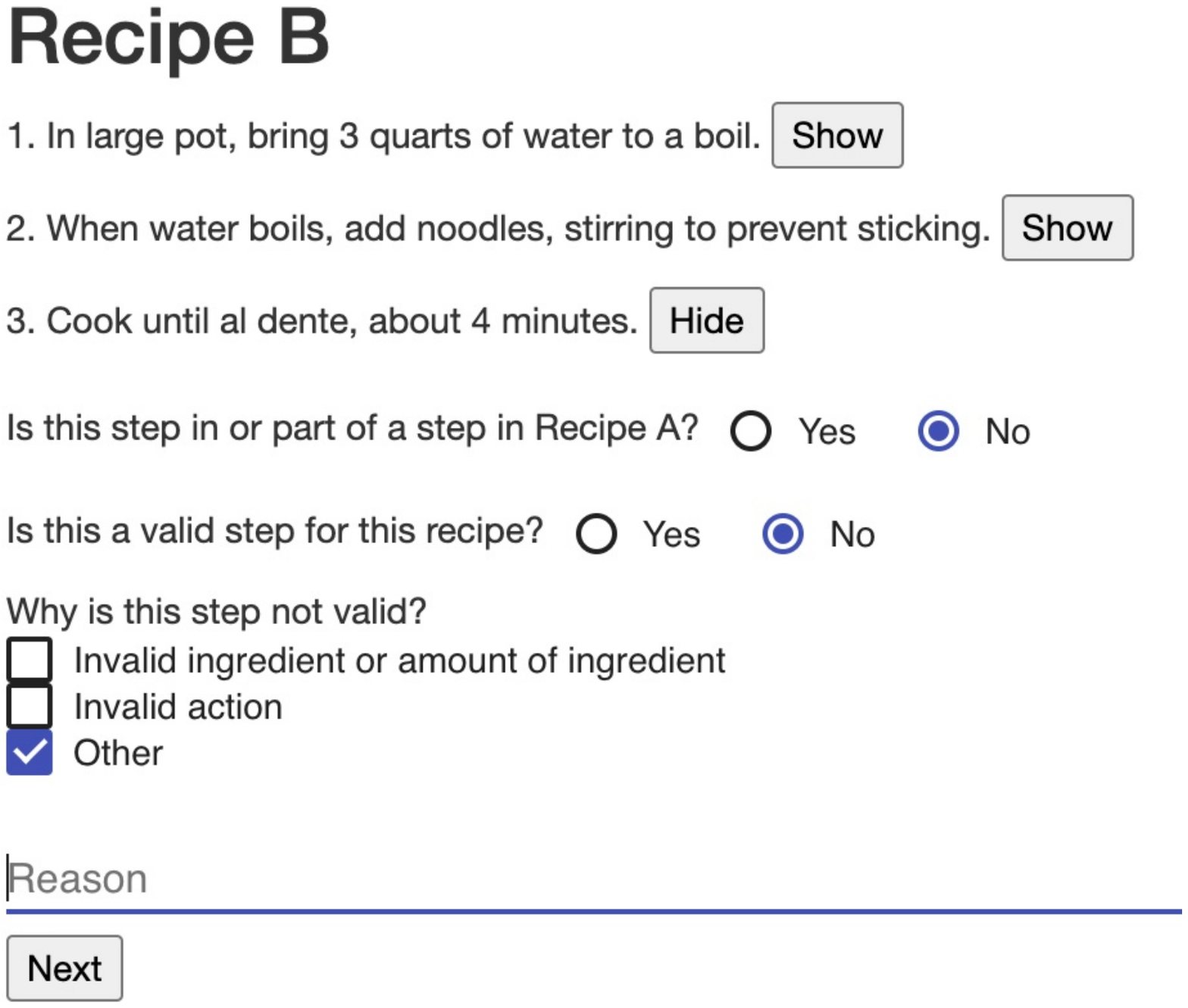}
    \caption{A Closer Look at the Step-By-Step Portion of Our MTurk Task. Annotators answer questions about Recipe B one step at a time. If the current step is not included in Recipe A, annotators are asked if it is a valid step. This distinguishes detrimental hallucinations from helpful elaborations. If the step is not valid, annotators are asked to indicate why using the checkboxes, which include a write-in ``other'' option annotators can also choose to hide or show individual steps for convenience.}
    \label{fig:mturk_invalid}
\end{figure}

Our work relies on human judgment to evaluate model performance, especially because we do not have reference information for revised recipes. We were also interested in measuring revisions that may be challenging to detect automatically: the addition of new information, the validity of new information, and the perceived usefulness of new recipes as a whole.

The most obvious human evaluation task would be to display original recipes and machine-revised recipes side-by-side and ask annotators to look for deletions, insertions, and other errors; we were concerned that the long inputs would make this style of task may be too strenuous for human annotators. Therefore, we designed our human evaluation task to decrease this strain. We opted to use Amazon Mechanical Turk because it offers a flexible platform to design evaluation interfaces.\footnote{We will release our HTML template to help future researchers design more sophisticated human evaluation task interfaces.} Screenshots of the interface are shown in Figures~\ref{fig:mturk_open}~and~\ref{fig:mturk_invalid}.

For each HIT, a annotator sees full set of ingredients and steps for a given recipe (Recipe A)---original or revised---on the left and center of the screen for a given dish. On the right, the other recipe (Recipe B) is shown one step at a time. We randomly flip if the original or revised recipe is Recipe A.\footnote{This was done to help mitigate the confounding of the order recipes are presented in and the comparison.} Annotators are asked if the current step of Recipe B is \textit{included} in Recipe A. If so, then they proceed directly to the next step. If not, we ask if the step is \textit{valid}, i.e. they would reasonably expect to perform this step while cooking the specified dish. If the step is furthermore not valid, we ask why: the step contains invalid ingredient, invalid action, or other (with a text input to explain why). These follow-up questions appear dynamically as well. When the annotator is finished with the current step, they hit the ``Next'' button to continue. Each step includes a button to show or hide the questions related to it for convenience.

When the annotator has reached the end of Recipe B, they answer three more questions: 1) their familiarity with the recipe so we can judge how much to trust their answers; 2) any steps in Recipe A that are missing in Recipe B; and 3) their preference for Recipe A, B, or neither. We ask annotators if a step from Recipe B is included in Recipe A \textit{and} if any steps from Recipe A are missing in Recipe B because we flip whether the original or revised recipe is static. We ask for their overall preference as a higher-level judgment of recipe quality.

\section{Qualitative Analysis}

\begin{figure}[ht]
    \centering
    \small
    \noindent
    \fbox{\parbox[t][][t]{.97\linewidth}{
        \textsf{{\normalsize Original Recipe} \\
        \textcolor{BrickRed}{1. Preheat oven to 350 degrees F.} \\
        \textcolor{orange}{2. Line baking sheets with parchment paper and set aside.} \\
        \textcolor{Dandelion}{3. In a medium bowl combine \textbf{first 5 ingredients (sugar - salt)}} \textcolor{ForestGreen}{ and then add the apples and toss.} \\
        \textcolor{blue}{4. Unroll the pie crusts on a floured board and cut 3-inch rounds \textit{(note in description)}, re-roll scraps and cut out more rounds \textbf{(need an} even number of rounds)}. \\
        \textcolor{Violet}{5. brush each round with beaten egg} \textcolor{Thistle}{and then place a large spoonful of apple mixture in center of half of the rounds,} \textcolor{brown}{top each with a plain round which you have cut a vent in, sealing edges with a fork or your fingers \textit{(we used fingers)},} \textcolor{SeaGreen}{sprinkle with sugar.} \\
        \textcolor{Fuchsia}{6. Bake pies on baking sheets for 20 minutes or until golden.}
    }}} 
    \fbox{\parbox[t][][t]{.97\linewidth}{
        \textsf{{\normalsize Revised Recipe} \\
        \textcolor{BrickRed}{1. Preheat oven to 350 degrees F.} \\
        \textcolor{orange}{2. Line baking sheets with parchment paper and set aside.} \\
        \textcolor{Dandelion}{3. In a medium bowl, whisk together \textbf{sugar, flour, nutmeg, cinnamon, and salt}.} \\
        \textcolor{ForestGreen}{4. Add the \textbf{peeled and diced} apples and toss to combine.} \\
        \textcolor{blue}{5. Unroll the pie crusts on a floured board and cut 3-inch rounds. Re-roll scraps and cut out more rounds \textbf{until you have} an even number of rounds.} \\
        \textcolor{Violet}{6. Brush each round with beaten egg.} \\
        \textcolor{Thistle}{7. Place a large spoonful of the apple mixture in the center of half of the rounds.} \\
        \textcolor{brown}{8. Top each filled round with a plain round and use a fork or fingers to seal the edges.} \\
        \textcolor{SeaGreen}{9. Sprinkle with sugar.} \\
        \textcolor{Fuchsia}{10. Bake pies on baking sheets for 20 minutes or until golden.}
        }}}
    \caption{Comparison between Original and Revised Recipes for ``Mini Apple Pies''. We first use 10 colors for the revised steps, then color semantically equivalent spans in the original recipe. Significant revisions between the recipes are \textbf{bolded}, while phrases that are deleted are \textit{italicized} in the original. }
    \label{fig:compare_recipes}
\end{figure}

\subsection{Case Study: Mini Apple Pies}
Figure~\ref{fig:compare_recipes} compares an original recipe and a revised recipe for ``Mini Apple Pies''\footnote{\url{https://www.food.com/recipe/mini-apple-pies-503243}}. We observe that the revisions are an overall improvement, and the steps are easier to follow. The revised recipe keeps the simple steps the same (1, 2, 6). As for minor edits, it deletes some author digressions (``we used fingers''), and makes a wording change (``need an even'' \textrightarrow ``until you have and even''). It also generates a helpful elaboration that the apples should be ``peeled and diced.''\footnote{The set of ingredients for this recipe actually specifies that the apples be peeled and diced; however, reiterating this information can still help end-users.}

The model makes two major revisions. First, it breaks the 5th original step, which is long and complex, into 4 separate revised steps. Second, it expands out the ``the first 5 ingredients'' from the 3rd original step into the 5 actual entities ``sugar, flour, nutmeg, cinnamon, and salt''. Both of these revisions demonstrate exactly the desired behavior, as end-users can more easily follow along, without having to remember the long text or looking back at the instructions (with flour-coated hands!). No hallucinations have occurred in this case study.

\section{Quantitative Analysis}

We collected human evaluations for 8 recipes,\footnote{We actually collected 10 recipes, but there was an unfortunate corruption for results from 2 recipes.} and got 3 annotations each (24 annotations total). The median time is 22.8 minutes, with a standard deviation of 33.7 minutes.\footnote{Based on pilot testing with our colleagues, we had estimated that each HIT would take 5 minutes to complete. We paid annotators \$1.20 per HIT (\~\$15 per hour), and will increase our rate in future annotation tasks to ensure fair compensation.} We believe that a few HITs were left idle for an extended period of time, distorting the average time to completion and standard deviation. This effect was also magnified by our small sample size. There are 12 unique workers, who completed an average of 2 annotations each.

We conduct a quantitative analysis of the human evaluation, reporting a suite of results.  The results are at either the  individual or aggregate levels. At the individual level, each annotation is a data point; this is applied to preferences and familiarity. At the aggregate level, the answers from all the annotators for that recipe go through a majority vote; this is applied to the other results.

\subsection{Aggregate-Level Analysis}
Recall that the reference recipe (Recipe A) can be either the `original' or the `generation' (i.e. the step-by-step recipe is the opposite). We separately analyze these conditions (4 recipes each) for selected results below.

\paragraph{Recipe Preferences}
We find across the 24 annotations, the generation is preferred 62.5\% of the time, the original 33.3\%, and neither 4.2\%. Interestingly, the propotions differ at the condition-level. When Recipe A is the generation, there is an 83.3\% generation preference, whereas when Recipe A is the original, there is a 58.3\% original preference. Future work should verify this bias towards preference Recipe A over more results.

\paragraph{Missing and Inclusion Steps}
First, we consider the condition where Recipe A is the original. We find that 86.0\% of (37/43) steps from the generation are included in Recipe B. Moreover, we find that 83.3\% (5/6) of the added steps are deemed valid -- which shows that the LLM was helpful most of the time. Annotators judged that 0\% of steps from the original were missing in the generation. 

Considering the condition where Recipe A is the generation, we find that 87.8\% (43/49) steps from the original are included in the generation. This is similar to the other condition. However, we find that only 33.3\% (2/6) of the steps that were ``added'' by the original are valid. This shows that the LLM's revision deleted several of the extraneous steps.\footnote{We also asked human annotators to classify \textit{why} a step is invalid, however, there was a bug that prevented it from working.}

Considering three-way Fleiss's kappa for all 8 recipes, we find moderate agreement for included steps ($\kappa=0.66$) and for valid steps ($\kappa=0.73$). This provides some empirical evidence that our instructions are clear, and our decomposition of the annotation work into several binary tasks is effective.

\subsection{Individual-Level Analysis}

\paragraph{Familiarity} At a scale of 1 to 5, where 1 is very unfamiliar with the recipe and 5 is very familiar with the recipe, the mean familiarity score is 2.58 and the median is 2.5. This coincides with our presumptions that revising recipes is hard since it requires domain and cultural knowledge to understand some of the recipes. 

We also conduct a Pearson Correlation test \cite{freedman2007statistics} between the levels of familiarity and the overall preference (i.e. original, LLM revised, or neither). We code the preference with numbers as follows: original (-1), neither (0), and LLM revised (1). The r-statistic for the correlation test is 0.0962 (p = 0.6547), showing that whether or not someone is familiar with the recipe does not have a correlation with the preferences.

\paragraph{Recipe Length} We also conduct a Pearson Correlation test \cite{freedman2007statistics} between the length of the recipe, both ``Recipe A'' and ``Recipe B,'' and the overall preference (i.e. original, LLM revised, or neither). We coded the preferences the same as in the case of familiarity. The r-statistic for the correlation test is 0.0613 (p = 0.776) for ``Recipe A'' and 0.1972 (p = 0.3556) for ``Recipe B.'' This shows that the length of either recipe does not correlate with the preference of the recipe.

\section{Conclusion and Future Work}
We have investigated the novel task of using LLMs to revise food recipes and improve their usability and quality. This domain is of particular interest because recipes are complex, diverse, and widely-used. Our work seeks to address how the machine reasoning capabilities of LLMs can assist real-world users. We first perform extensive experimentation with prompt engineering, and find the most success with an informative prompt which is grounded with the original recipe and ingredients list. We then prompted GPT-3.4 with a diverse sample of recipes. We perform both qualitative and quantitative analysis to verify the system's performance on our proposed task. Qualitatively, we designed a comprehensive human evaluation task which enables annotators to make granular comparisons between original and revised recipes. Highlights of our analyses include that annotators prefer the revised recipe 62.5\% of the time (vs. 33.3\% for original), and that 83.3\% of revised steps are deemed helpful. 

This work was conducted for the final project of CIS 7000-007 Machine Reasoning at the University of Pennsylvania, with the bulk of the time used for designing, pilot testing, and implementing the Amazon Mechanical Turk task. A direct extension of our work would be expanding our analysis to the full set of 300 diverse recipes. We also hope to investigate automated methods to evaluate the quality of revisions, and how well they correlate with the human evaluations. Beyond this particular work, we encourage further efforts in human evaluation design, model probing/interpretability, and human-centered applications of large language models.

\nocite{*}
\bibliography{references}
\bibliographystyle{acl_natbib}

\end{document}